\documentclass{article}

\usepackage[nonatbib,final]{neurips_2024}

\usepackage[utf8]{inputenc} 
\usepackage[T1]{fontenc}    
\usepackage{hyperref}       
\usepackage{url}            
\usepackage{booktabs}       
\usepackage{amsfonts}       
\usepackage{nicefrac}       
\usepackage{microtype}      
\usepackage{xcolor}         
\usepackage{amsmath}
\usepackage{adjustbox}
\usepackage{multirow}
\usepackage{wrapfig}
\usepackage{bbding}
\usepackage[ruled,linesnumbered]{algorithm2e}

\usepackage{expl3}
\ExplSyntaxOn
\newcommand\latinabbrev[1]{
  \peek_meaning:NTF . {
    #1\@}%
  { \peek_catcode:NTF a {
      #1.\@ }%
    {#1.\@}}}
\ExplSyntaxOff

\def\ie{\latinabbrev{i.e}}

\title{HumanVLA: Towards Vision-Language Directed Object Rearrangement by Physical Humanoid}

\author{%
Xinyu Xu$^{12*}$\hspace{2em} Yizheng Zhang$^{2*}$\hspace{2em} Yong-Lu Li$^{1}$\hspace{2em} Lei Han$^{2\dag}$\hspace{2em} Cewu Lu$^{1\dag}$\\
$^1$Shanghai Jiao Tong University \hspace{2cm} $^2$Tencent Robotics X\\
\small \texttt{ \{xuxinyu2000, yonglu\_li, lucewu\}@sjtu.edu.cn \{yizhenzhang, lxhan\}@tencent.com}
}

\begin{document}
\maketitle

\let\thefootnote\relax\footnotetext{$^*$ Equal contribution. $^\dag$ Equal advising.}
\let\thefootnote\relax\footnotetext{$^\ddag$ The code is available at \href{https://github.com/AllenXuuu/HumanVLA}{https://github.com/AllenXuuu/HumanVLA}.}

\vspace{-1cm}
\begin{figure}[h]
  \centering
  \includegraphics[width = 0.95\textwidth]{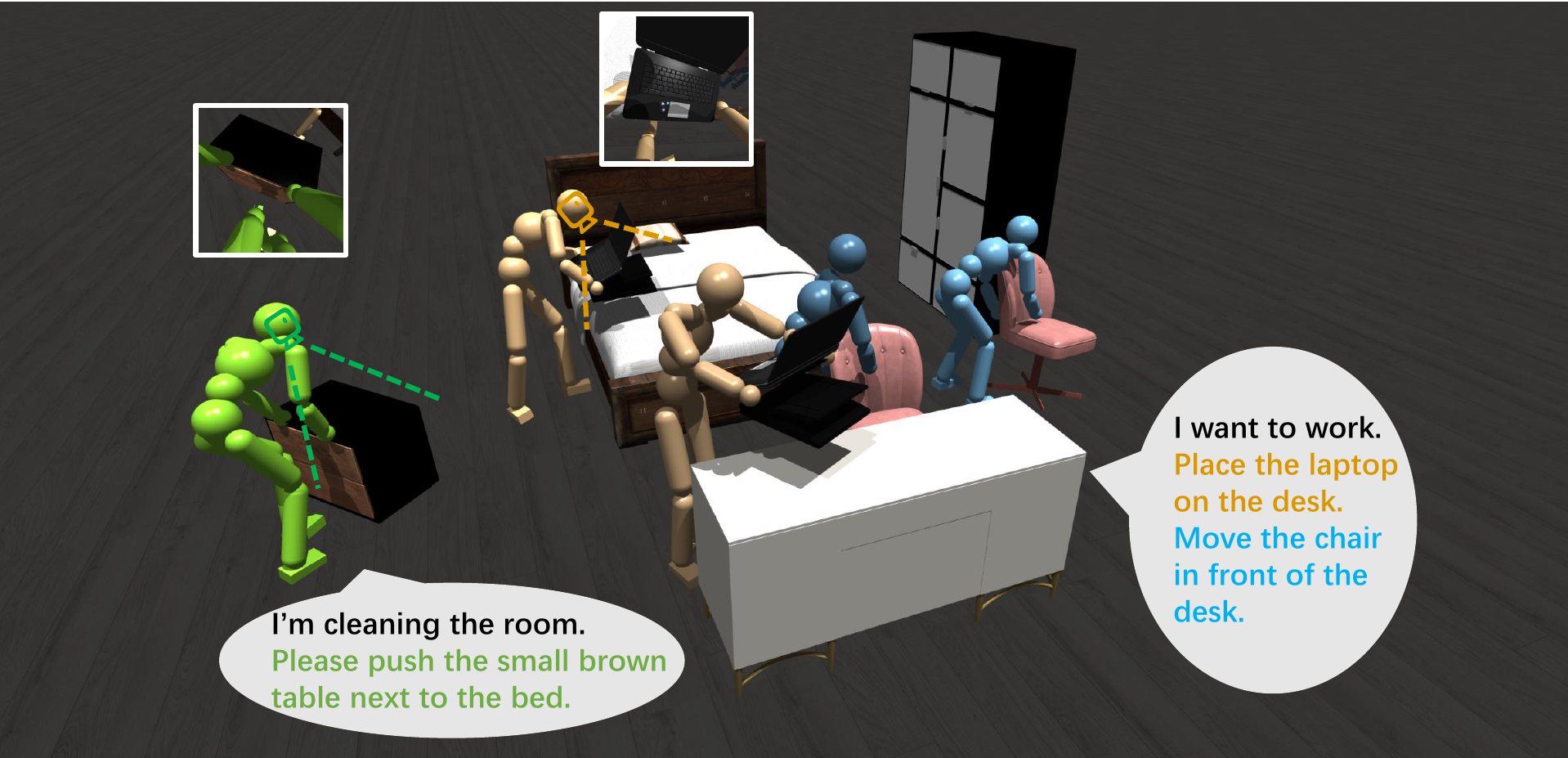}
  \caption{HumanVLA performs various object rearrangement tasks directed by the egocentric vision and natural language instructions.}
  \label{fig:teaser}
\end{figure}

\begin{abstract}
    Physical Human-Scene Interaction (HSI) plays a crucial role in numerous applications. 
    However, existing HSI techniques are limited to specific object dynamics and privileged information, which prevents the development of more comprehensive applications.
    To address this limitation, we introduce HumanVLA 
    for general object rearrangement directed by practical vision and language. 
    A teacher-student framework is utilized to develop HumanVLA.
    A state-based teacher policy is trained first using goal-conditioned reinforcement learning and adversarial motion prior.
    Then, it is distilled into a vision-language-action model via behavior cloning.
    We propose several key insights to facilitate the large -scale learning process.
    To support general object rearrangement by physical humanoid, we introduce a novel Human-in-the-Room dataset encompassing various rearrangement tasks.
    Through extensive experiments and analysis, we demonstrate the effectiveness of the proposed approach.
\end{abstract}
\section{Introduction}

Learning human-scene interaction (HSI) in realistic physical environments is a vital requirement of many applications, including computer graphics, embodied AI, and robotics. 
In this field, many previous efforts have been made to promote expressive humanoid control~\cite{amp,calm,phc,ncp}, static physical scene interaction~\cite{nsm,interphys,unihsi,interscene}, and manipulating a specific object~\cite{physhoi,interphys,xie2023hierarchical}.
These works have achieved great success in synthesizing plausible HSI controls.

Nonetheless, significant challenges persist in the realm of more extensive HSI applications and two primary issues need to be solved. 
\textit{Firstly}, the current techniques are limited to static objects, such as sitting on a chair~\cite{unihsi,interscene}, or specific object dynamics, such as carrying a box~\cite{interphys} and throwing a ball~\cite{physhoi}. 
However, in a more complicated real-world environment, humans demonstrate exceptional skills in manipulating a diverse range of objects with different geometries, poses, and weights.
It poses a challenging requirement on the varied dynamics of objects in HSI synthesis, \ie, a universal manipulation policy.
\textit{Secondly}, ground-truth object and goal states are necessary to direct humanoid controls in previous works.
However, without the help of external localization devices, this privileged information is difficult to access in a real-world transfer.
It prohibits practical real-world extensions like humanoid robots and necessitates an easily deployable perception method.

Our work takes a step forward in the above two challenges.
We investigate the concept of general-purpose object rearrangement performed by a physically interactive humanoid.
The whole-body physical humanoid is instructed to carry out daily object loco-manipulations in an indoor room setting.
The tasks involve human-like motion controls, interaction with diverse objects, and following desired object dynamics.
Moreover, considering the unavailability of privileged information about object and goal states in real-world humanoid applications, we delve into humanoid controls directed by practical vision and language. 
Compared to privileged states, vision-language modalities are more accessible and offer new potential for practical applications.  
It also presents an ultimate vision of the research community on humanoids: a human-like agent capable of understanding language instructions, perceiving its environment, and executing daily tasks to assist humans. 
Fig.~\ref{fig:teaser} provides intuitive examples of our work, where the humanoid agent can push a table, carry a laptop, and pull a chair, all directed by vision and language. 
Comparisons of our work with previous studies are available in Tab.~\ref{tab:compare}.

Our work starts with learning state-based teacher policy and then distills the policy into a vision-language-action model.
In the first stage, we train the policy using goal-conditioned reinforcement learning and adversarial motion priors (AMP)~\cite{amp}, within a generative adversarial imitation learning~\cite{gail} paradigm. 
The discrimination reward plus task-conditioned reward encourages humanoids to generate realistic motions and complete the task.
However, interacting with diverse objects remains challenging for vanilla AMP. 
We introduce improved techniques to facilitate general manipulation, in-context navigation, and prioritized task completion.
In the second stage, we distill the policy into a student network, named \textbf{HumanVLA}, an end-to-end vision-language-action model for physical humanoid.
Behavior cloning~\cite{bain1995framework} is used to train the student HumanVLA, \ie, cloning the teacher action at each step.
A challenge of learning VLA models is the poor perception quality of the unconstrained camera pose.
We propose a novel active rendering technique to improve gaze intention.

To support HumanVLA, we create a novel dataset named \textbf{Human-in-the-Room (HITR)}. 
It consists of four different room layouts: \textit{bedroom}, \textit{livingroom}, \textit{kitchen}, and \textit{warehouse}.
Each layout is populated with \textit{separated}, \textit{instantiable}, and \textit{replaceable} objects from HSSD~\cite{hssd} assets to create diverse scenes. 
The humanoid agent is placed in the scene with an instruction to rearrange the room.
Statistically, the HITR dataset consists of 50 static objects and 34 movable objects.
In our extensive experiments, we train HumanVLA in IsaacGym~\cite{isaacgym} with tasks from HITR.
Results demonstrate the effectiveness of our method in generalized object rearrangement and vision-language perception.

In summary, our contributions include: (1) We study general object rearrangement by physical humanoids. Several advanced techniques are introduced to interact with diverse objects. (2) We propose HumanVLA, the first vision-language-action model on physical humanoids to complete tasks directed by egocentric vision and natural language instruction. (3) We propose the HITR dataset to facilitate research in this field. Comprehensive experiments are conducted in HITR to validate the effectiveness of our method.

\begin{table}[!t]
\caption{Comparisons between HumanVLA and past works.}
\label{tab:compare}
\centering
\resizebox{0.95 \linewidth}{!}{
\begin{tabular}{l|ccccccc}
\toprule
\multirow{2}{*}{Methods}& \multirow{2}{*}{Physics} & Object  & Object    & Language & \multirow{2}{*}{Ego-Vision}& \# Static& \# Movable\\
                        &                          & Interaction & Dynamics  & Instruction &                          & Objects & Objects
\\
\midrule

NSM~\cite{nsm}             & & \checkmark & \checkmark & & & 25 & 2 \\
SAMP~\cite{samp}           & & \checkmark & & & & 7 & -\\
OMOMO~\cite{omomo}         & & \checkmark & \checkmark & \checkmark & & - & 19 \\
PADL~\cite{pald}           & \checkmark & & & \checkmark & & - & - \\
InterPhys~\cite{interphys} & \checkmark & \checkmark & \checkmark & & & 350 & 1 \\
InterScene~\cite{interscene} & \checkmark & \checkmark & & & & 57 & - \\
UniHSI~\cite{unihsi} & \checkmark & \checkmark &  & \checkmark & & 40 & -\\
\midrule
HumanVLA(Ours) & \checkmark & \checkmark & \checkmark & \checkmark & \checkmark & 50 & 34\\
\bottomrule
\end{tabular}

}
\vspace{-0.3cm}
\end{table}
\section{Related Works}

\textbf{Motion Synthesis} is a long-term research topic in graphics, vision, and robotics.
It can be divided into two streams: kinematic motion synthesis~\cite{nsm,starke2022deepphase,holden2017phase,samp,omomo,huang2023diffusion,behave,jiang2024scaling,dimos,motiongpt,motionx,circle,cai2021unified} and physics-based motion synthesis~\cite{deepmimic,amp,ase,pald,unihsi,phc,calm,ncp,humanoidbench,xie2023hierarchical,interscene,physhoi,braun2023physically,motionvae}.
Kinematic methods aim at synthesizing visually plausible motions with less penetration, floating, and being semantically faithful.
It leverages generative neural networks like VAEs~\cite{samp,cai2021unified}, Transformers~\cite{circle,motiongpt}, or Diffusions~\cite{omomo,huang2023diffusion,jiang2024scaling} to predict next state.
Our work belongs to the physics-based methods, which have an additional requirement on physical plausibility.
It follows a control-then-model paradigm where the control is typically achieved by a learning algorithm, and the model is constrained by a physics simulator.
DeepMimic~\cite{deepmimic} uses reinforcement learning plus imitation learning to track motion references and perform versatile motion controls.
NCP~\cite{ncp} advances motion tracking with discrete latent prior.
Adversarial Motion Prior (AMP)~\cite{amp} uses generative adversarial imitation learning to learn natural state transition from unstructured motion data.
It is further extended with a reusable controller~\cite{ase}, high-level language~\cite{pald}, expressive control~\cite{phc}, and latent conditions~\cite{calm}.
Recently, there has been an increasing emphasis on the synthesis of interactive motions.
InterPhys~\cite{interphys} uses task-conditioned reward plus stylized adversarial reward to perform HSI tasks such as sitting, lying, and box carrying.
InterScene~\cite{interscene} extends the paradigm to synthesize long-horizon static interactions.
UniHSI~\cite{unihsi} leverages the vast knowledge of the language model to provide a unified interface for static HSI.
However, previous works are limited to static objects or specific movable objects but fail to interact with various objects. 
In contrast, our research studies general object rearrangement in a daily room, posed with challenges in diverse object geometries, positions, and weights.

\textbf{Room Rearrangement} is a crucial application of embodied AI, where an instructed agent is placed in a room to search, navigate, and interact with desired objects.
Recent efforts~\cite{RoomR,habitat2,homerobot,ai2thor,procthor} have proposed various platforms and benchmarks to facilitate room rearrangement research.
Visual room rearrangement~\cite{RoomR} takes the agent to transverse both goal and initial scenes to recover object states based on visual observations.
OVMM~\cite{homerobot} presents open-vocabulary pick-and-place manipulation challenges in pursuit of extreme generalization capability.
Recent algorithms~\cite{wu2023tidybot,ding2023task} leverage commonsense knowledge in large language models to plan rearrangements.
However, these works are designed for simple embodiments, such as disc-shaped mobility and gripper manipulation.
They are limited to moving on smooth terrain and handling only small-sized objects.
In contrast, our work pioneers the design of rearrangement tasks in a complex human-like embodiment. 
It benefits from bipedal locomotion and stronger interaction motors.
For example, our model is capable of carrying 20\textit{kg} objects, which is beyond the capabilities of traditional stretches.

\textbf{Vision-Language-Action (VLA) Model} maps practical vision-language input to generate action controls.
It has demonstrated impressive results in the fields of embodied AI and robotics~\cite{palme,rt2,rtx,rth,zhen20243dvla}.
Thanks to the robust scalability of the vision and language modalities, VLAs also benefit from large-scale training~\cite{rt2,rtx}, opening up the potential for more general-purpose applications.
However, existing VLAs are designed for simple embodiments, such as desktop gripper manipulation.
The exploration of VLAs for more complex, high-dimensional humanoids is still in its early stages.
Our work is the first to develop humanoid controls directed by practical vision and language.
\begin{figure}[!t]
  \centering
  \includegraphics[width = 0.95\textwidth]{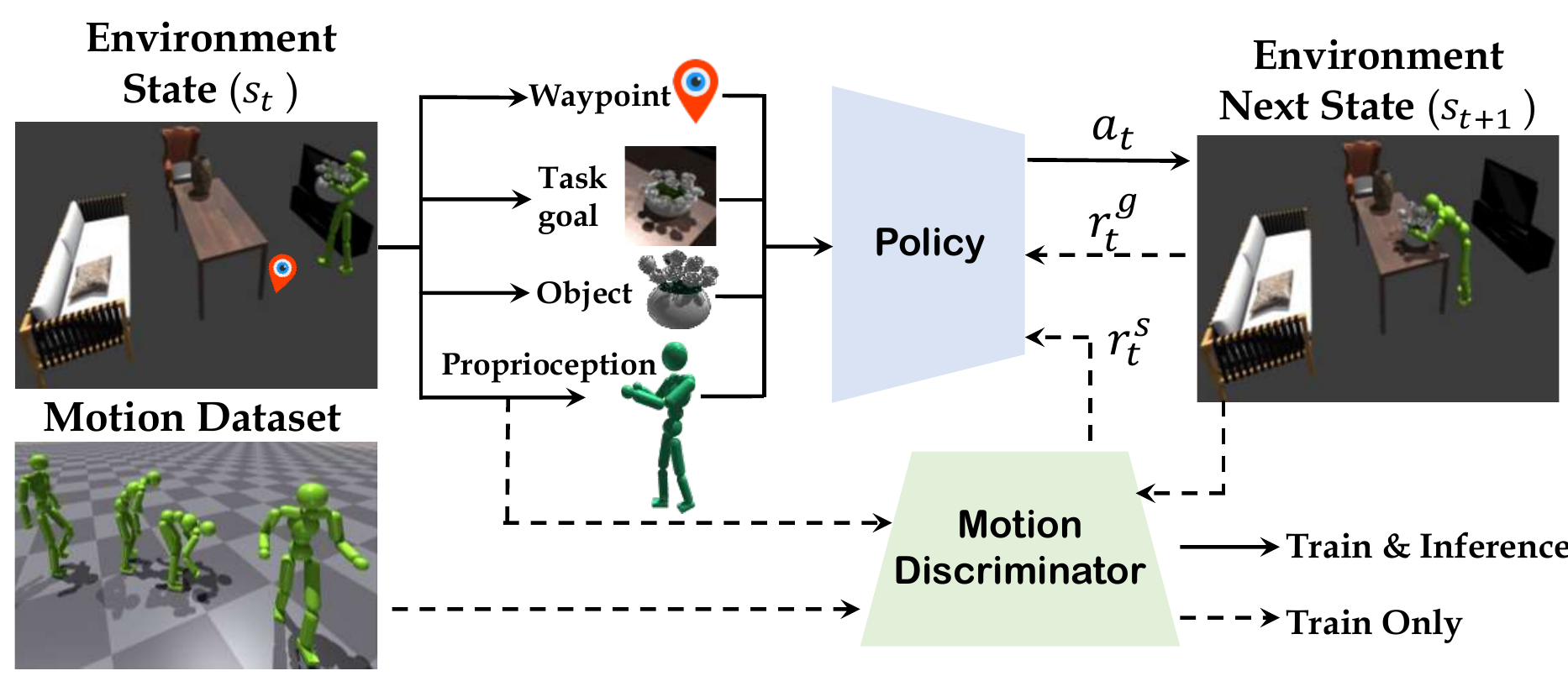}
  \caption{An overview of learning state-based HumanVLA-Teacher policy using goal-conditioned reinforcement learning and adversarial motion prior.}
  \vspace{-0.6cm}
  \label{fig:teacher}
\end{figure}

\section{Approach}
In this section, we introduce the learning process of HumanVLA.
Training HumanVLA directly through large-scale reinforcement learning (RL) presents significant challenges, including a high-dimensional action space, a composite state space, slow rendering speed, and other common issues associated with large-scale RL.
To this end, we utilize a teacher-student framework to train HumanVLA, which has been validated in applications like dexterous re-oreintation~\cite{chen2021general} and grasping~\cite{xu2023unidexgrasp}.
It consists of two phases. 
In the first phase (Sec.~\ref{sec:teacher}), we leverage goal-conditioned reinforcement learning and adversarial motion priors~\cite{amp} to train HumanVLA-Teacher.
It is presented with the oracle scene state, including precise object pose, geometry, navigation waypoint, and goal coordinate.
In the second phase (Sec.~\ref{sec:student}), we operate in a more practical setting, where the egocentric vision is tasked with perceiving the scene, and natural language instruction is used to specify the goal.
In a blueprint of real-world humanoid robots, observations used by HumanVLA are all accessible in a real-world deployment.
HumanVLA is trained via behavior cloning~\cite{bain1995framework} from HumanVLA-Teacher, where the pre-trained teacher policy significantly reduces the compute demands of the process.

\subsection{State-based Teacher Policy Learning}
\label{sec:teacher}
We train HumanVLA-Teacher with complete knowledge of the scene state to enable a variety of object rearrangement tasks.
The rearrangement task is formulated as a reinforcement learning process.
To elaborate, at each time step $t$, given state $s_t$ and goal $g$, HumanVLA-Teacher $\pi_{tch}$ predicts an action $a_t$ from policy distribution $\pi_{tch}(a_t|s_t,g)$.
The action $a_t$ is processed by a physics simulator $f(s_{t+1}|a_t,s_t)$ to generate the next state $s_{t+1}$.
The learning objective is to maximize the accumulated reward $\mathcal{R}(\pi_{tch})=\sum_{t=0}^{T-1}\gamma^t r_t$ where $\gamma$ is the discount factor and $r_t$ is the step reward at time $t$.

To train robust policies that enable humanoids to interact with objects and achieve various goals $g$ in a life-like manner, it is crucial for the humanoids to learn from authentic human motions and generalize across different tasks.
To this end, we use goal-conditioned task reward $r^G(g,s_t,s_{t+1})$ to encourage the agent to complete the task and style reward $r^S(s_{:t+1})$ to imitate human motion prior. 

We employ adversarial motion prior (AMP)~\cite{amp} to model the style reward, which incorporates an adversarial discriminator $D$ to discriminate motions from simulated synthesis or tracked motion dataset.
It is trained with the objective:
\begin{equation}
\label{eq:amp}
\begin{aligned}
    \operatorname*{argmin}_D 
    -E_{d^M(s_{t:t+t^*})}[\log(D(s_{t:t+t^*}))] 
    -E_{d^\pi(s_{t:t+t^*})}[\log(1-D(s_{t:t+t^*}))] \\
     + w^{gp}E_{d^M(s_{t:t+t^*})}\bigg[\Big|\Big|\nabla_\phi D(\phi)\Big|_{\phi=s_{t:t+t^*}}\Big|\Big|^2\bigg],
\end{aligned}
\end{equation}
where $d^M(s_{t:t+t^*})$ and $d^\pi(s_{t:t+t^*})$ are distributions of $t^*$-frame motion clips from dataset $M$ and policy $\pi$. The first two items in Eq.~\ref{eq:amp} are to discriminate motions while the last item with a coefficient $w^{gp}$ regularizes the gradient penalty~\cite{mescheder2018training} in adversarial training.
The style reward $r^S$ to encourage realistic motion synthesis is then formulated as
\begin{equation}
    \label{eq:style_reward}
    r^S(s_{:t+1}) = -\log(1 - D(s_{t+1-t^*:t+1})).
\end{equation}

We uniformly conceptualize goal-conditioned object rearrangement as three processes: locomotion towards the object, contacting the object, and relocating the object to the goal. 
These steps are accomplished by unified progressively increasing task rewards.

Despite the powers of goal-conditioned reinforcement learning and adversarial motion prior, generalized object rearrangement tasks by humanoids still pose significant challenges. 
Previous works~\cite{interphys,interscene,unihsi} have been limited to simple tasks such as static sitting, lying, or carrying a specific box.
We propose new techniques to overcome challenges in generalized object rearrangement.
Generalized object interaction involves geometrically various objects.
However, tracking motion data for each individual object is labor-intensive, and infeasible to tackle novel objects.
We expect RL to enable automatic object generalization.
Thus, we encode object geometry to learn a geometry-aware policy and design a carry curriculum to facilitate the learning.
Due to the misalignment of objects in human motion data and the task, we propose style reward clipping to prioritize high-level task execution.
Navigating in a complex room requires high-level planning to avoid collisions, we use in-context path planning to enable efficient locomotion.
More detailed explanations of our improved techniques are described in the following:

\textbf{Geometry Encoding.} 
Object state is crucial in HSI synthesis. 
Previous studies~\cite{interphys,physhoi} primarily encode object position, rotation, and linear and angular velocities to act on certain objects like boxes or balls.
A general policy for interactions with diverse objects should incorporate geometric information.
Thus, we augment the teacher policy with geometric object representations via Basis Point Set (BPS)~\cite{bps} encoding.
A shared set of basis points is randomly sampled from a unit sphere and encodes object geometry using delta vectors from each basis point to the nearest object point.
In contrast to geometries encoded by a neural net~\cite{qi2017pointnet}, BPS encoding is computationally efficient and accelerates policy learning.
Consequently, we use object geometry, position, rotation, and linear and angular velocities to form a comprehensive object observation, thereby facilitating a more expressive policy control.

\textbf{Carry Curriculum Pre-training.} 
Object rearrangement is conceptualized as a three-step process in the aforementioned paragraph.
However, directly learning the entire three-step rearrangement task from scratch is challenging due to the long task horizon.
Besides, physics-based object movement presents greater challenges compared to kinematic object movement~\cite{omomo}, primarily because the object state is not directly editable. 
Instead, it requires indirect control of the physical humanoid to interact.
To this end, we draw inspiration from the curriculum learning ~\cite{bengio2009curriculum} and design an easy carry curriculum to pre-train the policy.
The carry curriculum only includes the first two of three steps: locomotion towards the object and carrying up the object for an in-the-air holding.
The carry curriculum has a shorter horizon and is empirically easier to converge.
Furthermore, the pre-trained in-the-air carry prior significantly benefits the subsequent object relocation.
For the carry curriculum, we use objects excluding those on the ground, such as tables and chairs, which are easier to move by pushing and pulling along the ground without a lift.
The carry curriculum shares the same learning paradigm with the rearrangement task, except for a different two-stage reward design.

\textbf{Style Reward Clipping.}
General object rearrangement involves manipulating novel objects that are not recorded in tracked motion data.
This creates a misalignment in optimization directions: strictly imitating reference motion or ensuring high-level task execution.
Previous work~\cite{interphys} balanced two items by a weighted sum between task reward and style reward in motion-aligned tasks.
However, in our general object rearrangement setting, goal-conditioned task exploration progress can be stagnant and the policy may learn actions devoid of task semantics following the logarithmic gradient in Eq.~\ref{eq:style_reward}.
For instance, when the object is difficult to lift, the policy tends to mimic insignificant hand swings in the motion data, rather than exploring carry-up actions.
We insert a style reward clipping to prioritize task execution, formulated as follows:
\begin{gather}
    \xi_t = \max(r^G(g,s_t,s_{t+1}), \xi_{min}), \\ 
    r_t = w^G r^G(g,s_t,s_{t+1}) + w^S \min( r^S(s_{:t+1}), \xi_t ),
\end{gather}
where $w^G,w^S$ are coefficients, and $\xi_t$ is the upper bound for the style reward. 
This formulation prioritizes goal-conditioned task execution over motion imitation in reward maximization.
In addition, we use a minimum upper bound $\xi_{min}$, to ensure basic motion imitation during the early stages when the task reward is near zero.

\textbf{In-context Path Planning.} 
Navigating a populated room requires high-level knowledge since the dense object cluster may collide with the humanoid agent and obstruct natural locomotion.
We use in-context path planning to guide the navigation.
Point clouds of all objects in the scene make up the spatial occupancy.
We perform top-down point projection and grid discretization to derive a 2D obstacle map of $20cm$ x $20cm$ grids.
We then plan a navigable path from the starting position to the object, and subsequently to the goal using $A^*$ algorithm~\cite{hart1968formal}, represented as a series of navigation waypoints to guide locomotion at each step.

Incorporating all the above features, we present an overview of training HumanVLA-Teacher in Fig.~\ref{fig:teacher}. 
Navigation waypoint, task goal, object state, and humanoid proprioception are sent to the policy network to derive an action.
The learning process is guided by task reward and motion discrimination reward.
Further details about the learning process can be found in the appendix.

\begin{figure}[!t]
  \centering
  \includegraphics[width = \textwidth]{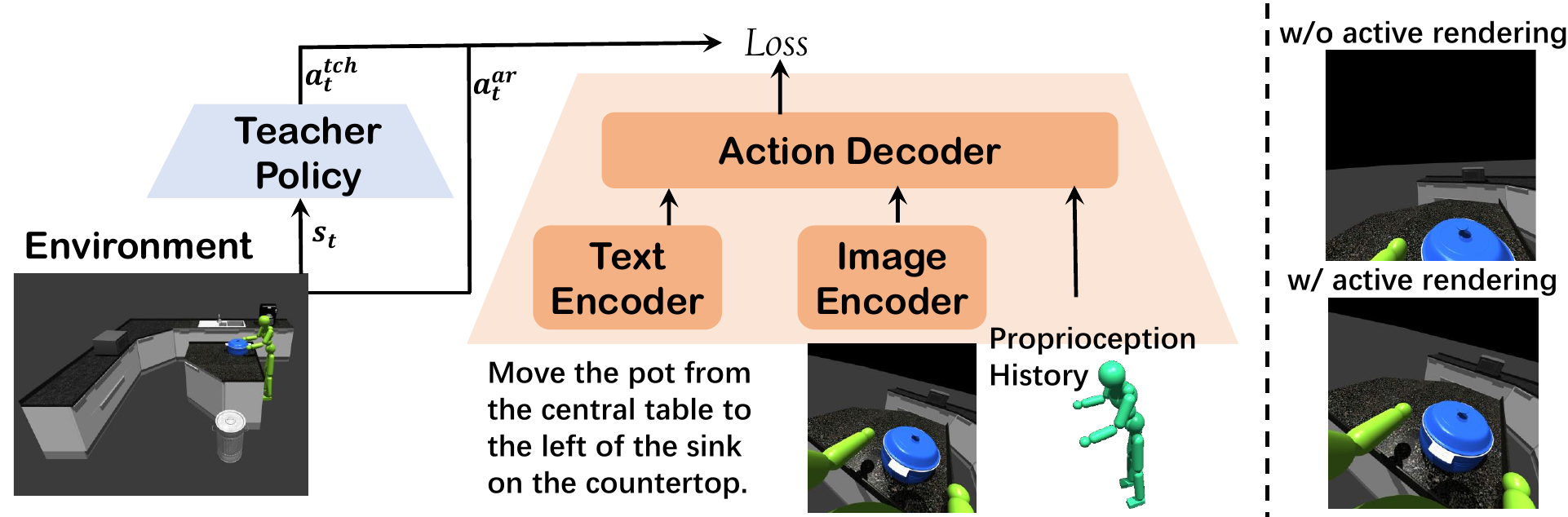}
  \caption{
  \textbf{Left}: An overview of learning HumanVLA by mimicking teacher action and active rendering action.
  \textbf{Right}: Comparison between \textit{w/} and \textit{w/o} active rendering. Active rendering leads to a more informative perception of human-object relationships.
  }
  \vspace{-0.6cm}
  \label{fig:student}
\end{figure}

\subsection{Distilling into Vision-Language-Action Model}
\label{sec:student}
While the HumanVLA-Teacher $\pi_{tch}$ leverages privileged information such as object state, goal state, and waypoint in the global coordinates, our biggest goal is towards practical humanoid control free of privileged information and real-world deployable. 
To this end, we replace privileged states with flexible egocentric vision and natural language instruction.
Notably, the proprioception observation is the only kept item from HumanVLA-Teacher to HumanVLA, which is represented in the local coordinate and can be obtained via forward kinematics and temporal differentiation.
The history action is also used in observation.
To obtain egocentric vision, we mount a mobile camera on the head of the human. 
It renders a 256 x 256 image with a field of view spanning 90 degrees at each step.
An overview of training HumanVLA is illustrated in Fig.~\ref{fig:student}.

\textbf{Behavior Cloning.} 
We train HumanVLA $\pi_{vla}$ using a teacher-student framework to distill the knowledge from HumanVLA-Teacher via behavior cloning~\cite{bain1995framework}. 
HumanVLA employs an \textit{EfficientNet-B0}~\cite{efficientnet} for image encoding and a frozen \textit{bert-base-uncased}~\cite{bert} for language encoding, whose features, along with proprioception, last action, are sent to the action decoder to derive an action.
At each time step $t$, HumanVLA-Teacher leverages privileged state $s_t$ and $g$ to derive an action $\pi_{tch}(a_t|s_t,g)$ while HumanVLA derives an action $\pi_{vla}(a_t|p_t,a_{t-1},v_t,l)$ based on proprioception $p_t$, last action $a_{t-1}$, egocentric image $v_t$, and language instruction $l$.
Behavior cloning bridges distributions between $\pi_{vla}(a_t|p_t,a_{t-1},v_t,l)$ and $\pi_{tch}(a_t|s_t,g)$, which can be directly implemented via supervised learning.
In the empirical training process, we observe a severe covariate shift problem in offline behavior cloning.
Thus, we opt for a DAgger~\cite{dagger} framework to train HumanVLA which alleviates the problem via online learning.

\textbf{Active Rendering.}
Though HumanVLA-Teacher possesses comprehensive knowledge in versatile control, naive policy distillation still suffers from the gap of observation expressiveness.
For instance, while egocentric vision is used to perceive objects, the humanoid gaze might not properly focus on the target object but renders a less informative image of the background.
Consequently, the perception quality of HumanVLA is significantly affected by the camera pose.
However, an optimal camera pose, determined by the head pose, is not guaranteed in the teacher policy, which only imitates a life-like head motion but ignores the vision quality.
We propose an active rendering technique to encourage the camera to focus on the object.
We analytically calculate the head-to-object direction in the global coordinate and then derive a head orientation.
Inverse kinematics is used to obtain active rendering actions $a^{ar}_t$ for the neck joint. 
It is used to derive a mixed supervision $a^{vla}_t$ in conjunction with the teacher action $a^{tch}_t$, formulated as
\begin{equation}
    a^{vla}_t = (1-w^{ar}) a^{tch}_t + w^{ar} a^{ar}_t,
\end{equation}
where $w^{ar}$ is the coefficient for active rendering.
Notably, this is only applied to the neck joint, while other joints only follow the teacher action.


\section{Human-in-the-Room Dataset}

Existing datasets~\cite{RoomR,homerobot} for object rearrangement are primarily designed for stretches with disc-shaped mobility and gripper manipulation.
Human-like embodiment has different physical attributes, such as stronger motors to handle large furniture like chairs and tables.
Besides, we follow~\cite{amp,interphys,interscene,unihsi} to use a humanoid model with spherical hands, which can struggle with manipulating small-sized objects, such as picking up a towel.
To address these issues, we introduce a novel Human-in-the-Room (HITR) dataset, designed to facilitate vision-language directed object rearrangement tasks on a humanoid.
The HITR dataset includes carefully designed objects of various sizes, ranging from 21\textit{cm} to 126\textit{cm}, and provides a variety of rearrangement tasks in various rooms. 
Each task involves separated, instantiable, and replaceable objects with defined initial and goal states. 
Additionally, each task is accompanied by a natural language instruction generated by a Large Language Model (LLM).

In constructing the HITR dataset, we reference common objects used in room designs from~\cite{ai2thor,habitat2,homerobot} and utilize object models from HSSD~\cite{hssd} to create basic assets.
Object assets are manually resized to ensure the interaction friendliness.
We adopt the procedural generation pipeline from~\cite{procthor} to generate diverse scenes.
First, we manually design four room layouts: \textit{bedroom}, \textit{livingroom}, \textit{kitchen}, and \textit{warehouse}, then randomly populate replaceable objects within these layout templates to establish the scene, as well as the goal state.
Next, we randomly relocate an object in the scene, either to the ground or another receptacle.
This relocated object is what the physical humanoid is tasked to rearrange.
We concatenate two rendered images of the initial and goal scenes and use the composite image to prompt \textit{gpt-4-vision}~\cite{gpt4} to generate an instruction.
The LLM is asked to distinguish between two states and provide an instruction to guide the state transition.
However, the LLM still struggles with understanding complex spatial relationships, such as left-right errors.
To ensure the quality of instructions, we manually review and revise them as necessary.
Ultimately, we build the HITR dataset of 615 tasks, with an average of 6.5 objects per task. 
There are 50 static objects like \textit{bed} and \textit{countertop}, as well as 34 movable objects like \textit{pillow} and \textit{vase}.
More details are in the appendix.

\section{Experiments}

\begin{table}[!t]
\caption{Results in box rearrangement. $\dag$ denotes our implementation.}
\label{tab:box}
\centering
\resizebox{0.9 \linewidth}{!}{
\begin{tabular}{l|ccc}

\toprule
& Success Rate (\%) $\uparrow$ & Precision (\textit{cm})$\downarrow$ & Execution Time (\textit{s}) $\downarrow$
\\

\midrule

InterPhys~\cite{interphys} & 94.3 & 8.3 & 9.1\\
InterPhys~\cite{interphys} $\dag$ & 97.8 & 12.6 & 5.3\\
HumanVLA-Teacher & \textbf{98.1} & \textbf{4.2} & \textbf{4.6}\\
\bottomrule
\end{tabular}

}
\vspace{-0.3cm}
\end{table}

\subsection{Settings}

\textbf{Datasets.} 
Our experiments are conducted on the HITR dataset. 
It is split into \textit{train} and \textit{test} subsets at a ratio of 9:1, containing 552 and 63 tasks respectively. 
The \textit{test} subset is used to evaluate the generalizability of our method in unseen tasks.
For the motion dataset used in training, we utilize OMOMO~\cite{omomo} and a locomotion subset from SAMP~\cite{samp}. 
OMOMO provides a variety of short-range motions involving moving different objects, while locomotion motions from SAMP enhance the locomotion aspect of our dataset.
We use 30-minute motions in total.
The source motion dataset features object rearrangement involving only seven different objects, which is far less than those in the HITR dataset. 
Despite this, we anticipate that our method can generalize to different objects.

\textbf{Metrics.} We adhere to a 10-second running time limit and follow~\cite{interphys} to evaluate methods with three metrics. 
\textbf{(1) Success Rate:} the proportion of tasks that are successfully rearranged within an error margin of $\theta$.
\textbf{(2) Precision:} the distance of the final object position to the goal.
\textbf{(3) Execution Time} the average time taken to complete a run.
For the Success Rate, a higher value indicates better performance, but for Precision and Execution Time, the lower the better.
All experiments are evaluated using 10 repeat runs.
For state-based methods, we follow~\cite{interphys} to set $\theta=20$\textit{cm}.
For vision-language-based methods, where the goal is specified via coarse instructions rather than precision goal coordinates, we set $\theta=40$\textit{cm}.
This criteria relaxation is also adopted in past works~\cite{chen2021general} for evaluating policies with different observations.

\subsection{Implementation Details}
We conduct experiments in parallel environments simulated using IsaacGym~\cite{isaacgym}, with neural networks implemented via PyTorch. 
Our physical humanoid model, following previous works~\cite{interphys,interscene,unihsi}, comprises 15 rigid bodies and 28 joints, each actuated by a PD-Controller.
The simulator runs at 60Hz, and the policy is queried at 30Hz.
The teacher policy is optimized using Proximal Policy Optimization~\cite{ppo} and takes two days on eight Tesla V100 GPUs to converge.
The student policy is trained using DAgger~\cite{dagger} and takes one day on two GPUs.
We provide comprehensive details about hyperparameters, neural architectures, observation space, and more in the appendix.

\subsection{Comparisons in Box Loco-Manipulation}
While our work is pioneering in the exploration of vision-language-directed general object rearrangement, direct comparisons with previous studies are difficult. 
The work most similar to ours is InterPhys~\cite{interphys}, which delved into state-based box loco-manipulation.
Due to the unavailability of training data, motions, and codes of InterPhys~\cite{interphys}, we instead refer to a box rearrangement subset in HITR to conduct experiments.
We train a state-based HumanVLA-Teacher using only box rearrangement tasks, along with an implementation of the InterPhys baseline.
Results of box rearrangement are reported in Tab.~\ref{tab:box}.
Given that the box is the simplest object to interact with, both methods achieve high success rates.
However, our method exhibits superior precision, with a result of 4.2 \textit{cm}, outperforming both the 8.3 \textit{cm} in the official report~\cite{interphys} and the 12.6~\textit{cm} in our implementation.
We use the standard deviation to evaluate the statistical significance of HumanVLA-Teacher in 10 repeated runs. 
The values are 0.02, 0.004, and 0.04 for Success Rate, Precision, and Execution Time respectively.
With high task completion rates and low variance, we demonstrate the effectiveness and robustness of our method in this first trial.

\begin{table}[!t]
\caption{Ablation study.}
\label{tab:abl}
\centering
\resizebox{0.9 \linewidth}{!}{
\begin{tabular}{lc|ccc}
\toprule
& Privileged State  & Success Rate (\%) $\uparrow$ & Precision (\textit{cm})$\downarrow$ & Execution Time (\textit{s}) $\downarrow$\\
\midrule
HumanVLA-Teacher               &\CheckmarkBold & {\bf 85.9} & {\bf 14.4} & 4.5 \\
\textit{w/o} geometry encoding &\CheckmarkBold & 64.5 & 43.4 & 5.5 \\
\textit{w/o} carry curriculum  &\CheckmarkBold & 66.3 & 73.4 & 5.3\\
\textit{w/o} style clipping    &\CheckmarkBold & 79.9 & 27.5 & {\bf 4.3}\\
\textit{w/o} path planning     &\CheckmarkBold & 67.8 & 37.2 & 5.5\\
\midrule
HumanVLA                       &\XSolidBrush   & {\bf 74.8} & {\bf 42.6} & {\bf 5.1}\\
\textit{w/o} active rendering  &\XSolidBrush   & 67.9 & 55.6 & 5.6\\
\textit{w/o} online learning   &\XSolidBrush   & 15.3 & 145.0& 8.2\\
\bottomrule
\end{tabular}

}
\vspace{-0.3cm}
\end{table}

\begin{table}[!t]
\caption{Results in unseen tasks.}
\label{tab:unseen}
\centering
\resizebox{0.9 \linewidth}{!}{
\begin{tabular}{lc|ccc}
\toprule
& Privileged State  & Success Rate (\%) $\uparrow$ & Precision (\textit{cm})$\downarrow$ & Execution Time (\textit{s}) $\downarrow$\\
\midrule
InterPhys~\cite{interphys}        &\CheckmarkBold  & 59.5 & 52.5 & 5.9 \\
HumanVLA-Teacher                         &\CheckmarkBold  & {\bf 79.3} & {\bf 19.3} & {\bf 4.6}\\
\midrule
Offline GC-BC~\cite{offlinebc}   &\XSolidBrush    & 10.2 & 152.3 & 8.5 \\
HumanVLA                                 &\XSolidBrush    & {\bf 60.2} & {\bf 57.0} & {\bf 5.8}\\
~~ \textit{w/o} active rendering         &\XSolidBrush    & 56.7 & 65.5 & 5.9\\
\bottomrule
\end{tabular}

}
\vspace{-0.3cm}
\end{table}

\subsection{Ablation Study}

We conduct comprehensive ablation studies on the \textit{train} split to validate each design choice, with results presented in Tab.~\ref{tab:abl}. 
Firstly, we evaluate the impact of improved techniques in HumanVLA-Teacher training, which achieves a success rate of 85.9\% and a precision of 14.4\textit{cm}. 
However, eliminating any component leads to a decline in performance.
The inclusion of geometry encoding and carry curriculum enables the model to manipulate a variety of objects effectively. 
Without either of these components, the success rate experiences a drop of approximately 20\%.
Style reward clipping prioritizes task execution, whose absence results in a 6\% decrease in the success rate.
Path planning helps humans navigate complex scenes. Its removal leads to a significant 18.5\% decrease in the success rate.
Subsequently, we validate design choices in training HumanVLA directed by vision and language. 
The default HumanVLA achieves a success rate of 74.8\% with a precision of 42.6\textit{cm}.
However, the absence of active rendering results in a substantial 6.9\% success rate drop, emphasizing the importance of perception quality.
We implement an offline behavior cloning baseline using ten off-the-shell teacher trajectories per task for training.
Without online learning, the system suffers from a severe covariate shift and performs poorly.

\subsection{Generalizing to Unseen Tasks}
We use the \textit{test} split of HITR to evaluate the generalizability of methods.
The unseen data, which includes novel scene compositions, visual appearances, and language instructions, poses a significant challenge to our method.
Results are presented in Tab.~\ref{tab:unseen}.
The state-based HumanVLA-Teacher tends to be more robust in unseen data.
Relatively small drops in success rate (6.6\%) and precision (4.9\textit{cm}) demonstrate strong generalizability of RL when using privileged information.
Moreover, it consistently outperforms the InterPhys~\cite{interphys} baseline on all metrics.
Applying HumanVLA to unseen tasks turns out to be more challenging due to the complexity of vision and language modalities.
The success rate of HumanVLA decreases to 60.2\%, and the precision drops to 57.0\textit{cm}. 
However, HumanVLA still consistently outperforms baselines without active rendering and the offline goal-conditioned behavior-cloning (Offline GC-BC)~\cite{offlinebc} method.

\begin{figure}[t]  
    \centering
    \includegraphics[width=0.95\textwidth]{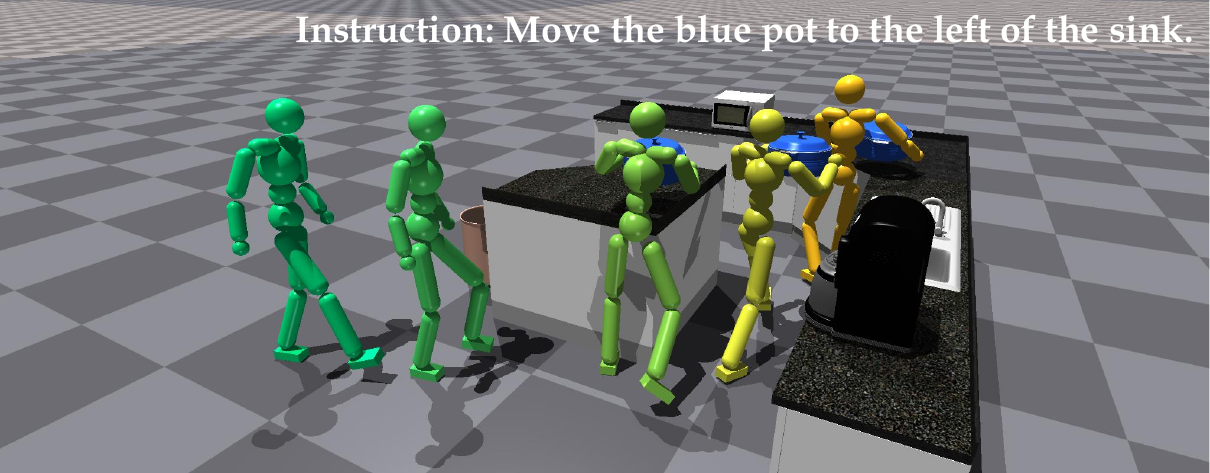}
    \includegraphics[width=0.95\textwidth]{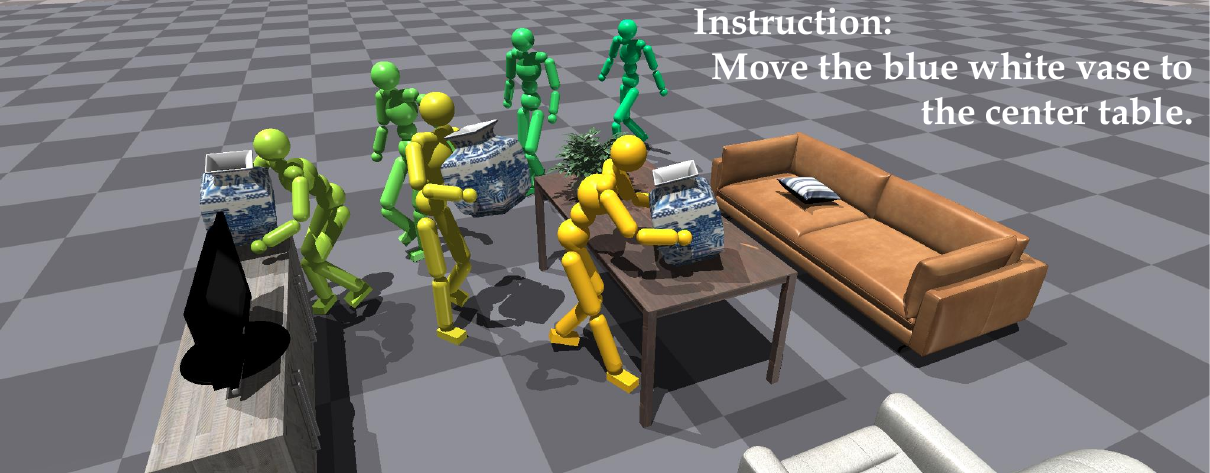}
    \caption{Qualitative results. The color transitions from green to yellow as the task progresses.}
    \label{fig:qualitative}
\end{figure}

\subsection{Qualitative Results}
We provide qualitative visualizations of how HumanVLA performs object rearrangement tasks in Fig.~\ref{fig:qualitative}. 
More results are available in the appendix. We demonstrate that HumanVLA is capable of moving a pot, vase, chair, and box based on language instructions.

\section{Conclusion}
We investigate vision-language directed object rearrangement by physical humanoids in this work, a fundamental technique for HSI synthesis and real-world humanoid robots.
Our system is developed using a teacher-student distillation framework.
We propose key insights to facilitate teacher policy learning with privileged states and introduce a novel active perception technique to favor vision-language-action model learning.
We present a novel HITR dataset to support our task.
In extensive experiments, our HumanVLA model demonstrates superior results in both quantitative and qualitative evaluations.
Future works include dexterous manipulation by physical humanoids and long-horizon multi-object interaction.

\section*{Acknowledgments}
This work was supported by the National Key Research and Development Project of China (No. 2022ZD0160102), National Key Research and Development Project of China (No. 2021ZD0110704), Shanghai Artificial Intelligence Laboratory, and XPLORER PRIZE grants.
\medskip

{\small
\bibliographystyle{plain}
\bibliography{ref}
}

\newpage
\appendix

\section{Limitations}
This work inherits the humanoid model from~\cite{amp,ase,interphys,interscene,unihsi} with spherical hands. It is hard to manipulate small-sized objects. Dexterous hands can be equipped to facilitate object manipulation in future works.

As the first work on general object rearrangement, our task settings only include one object movement at each time.
Long-horizon object rearrangement is left for future work.

The current version of our system does not contain explicit memorizing, planning, navigation, and multi-agent collaboration modules. We leave more ad-hoc designs to future work.

\section{Broader Impacts}
We study simulated physical humans in the work, whose technique holds the potential for extension to real-world humanoid robots.
This could have a significant positive societal impact, as humanoid robots have the potential to assist humanity in various ways.
However, it is crucial to carefully consider safety concerns associated with the use of humanoid robots.

\section{Licenses}
We use assets from ASE~\cite{amp}, HSSD~\cite{hssd}, OMOMO~\cite{omomo}, and SAMP~\cite{samp} in this work.
ASE is released under the NVIDIA license.
HSSD is released under the CC BY-NC 4.0 license.
OMOMO does not have a specified license.
SAMP is released with its license on its GitHub repository\footnote{\href{https://github.com/mohamedhassanmus/SAMP?tab=readme-ov-file\#license}{https://github.com/mohamedhassanmus/SAMP?tab=readme-ov-file\#license}}.

\section{Dataset}

\begin{figure}[t]
    \centering
    \includegraphics[width=\textwidth]{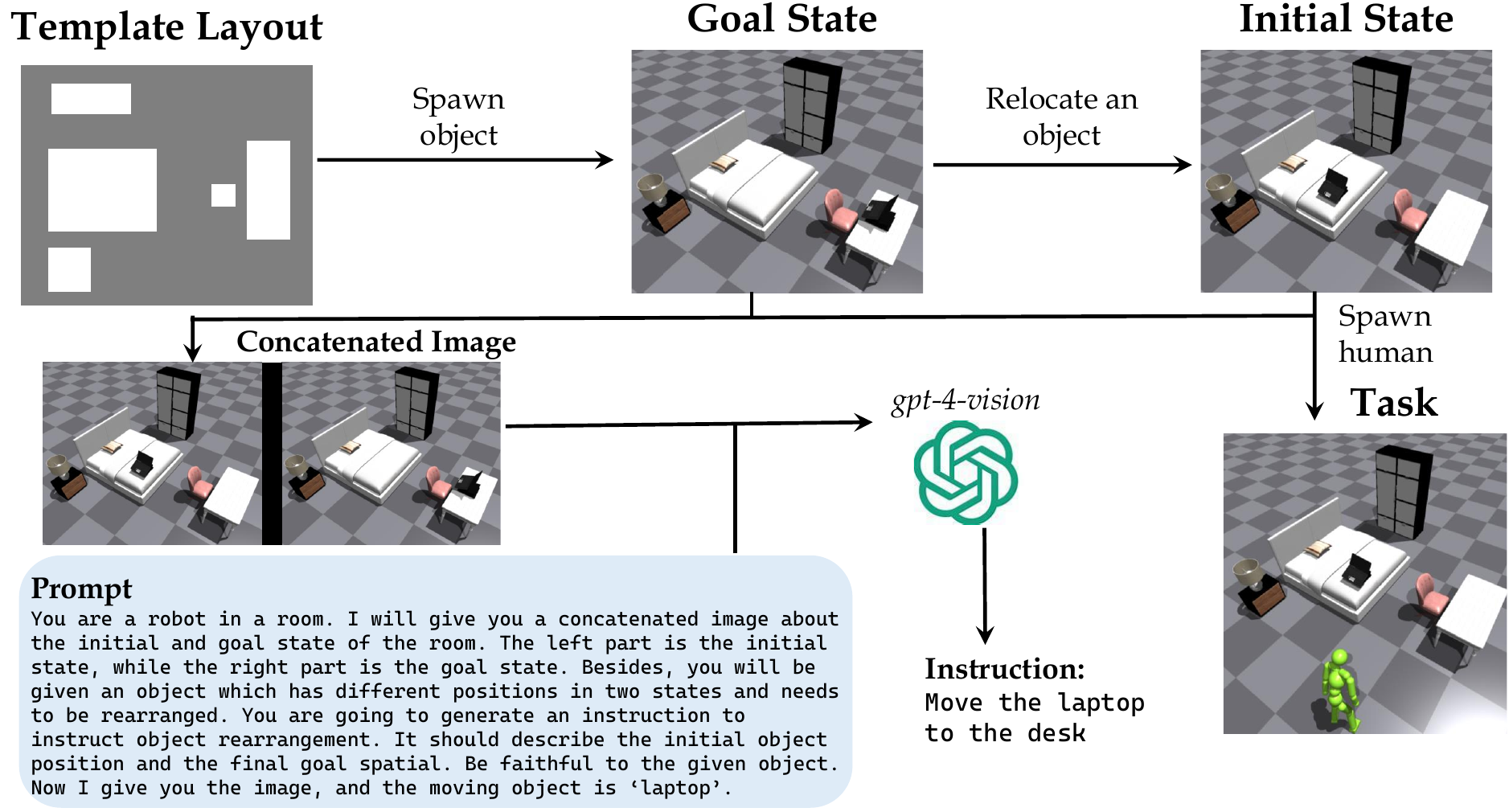}
    \caption{The task generation process of HITR dataset.}
    \label{fig:datagen}
\end{figure}
\begin{figure}[t]
    \centering
    \includegraphics[width=.7\textwidth]{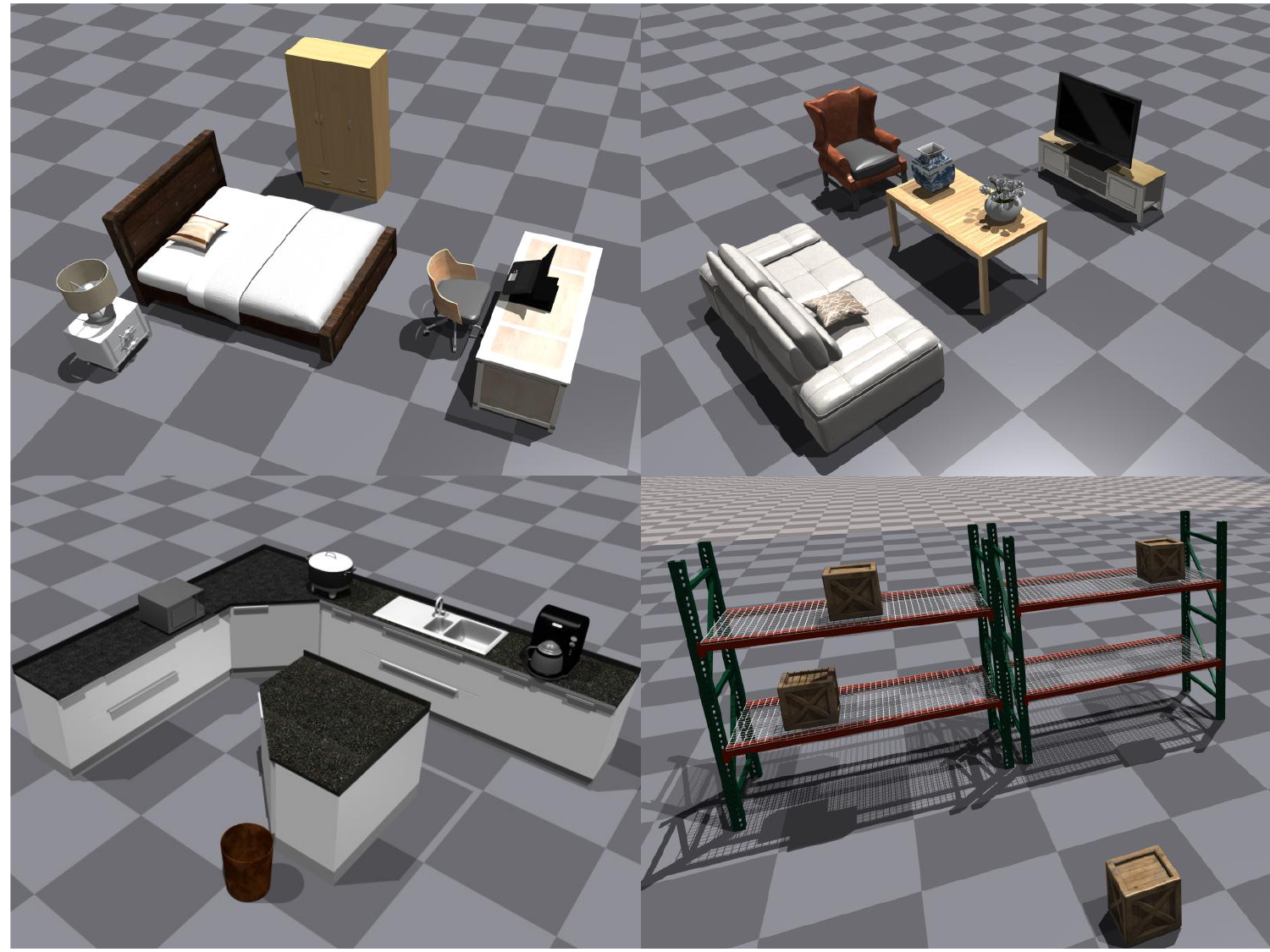}
    \caption{Different rooms in HITR dataset.}
    \label{fig:hitr-room}
\end{figure}
\begin{figure}[t]
    \centering
    \includegraphics[width=.7\textwidth]{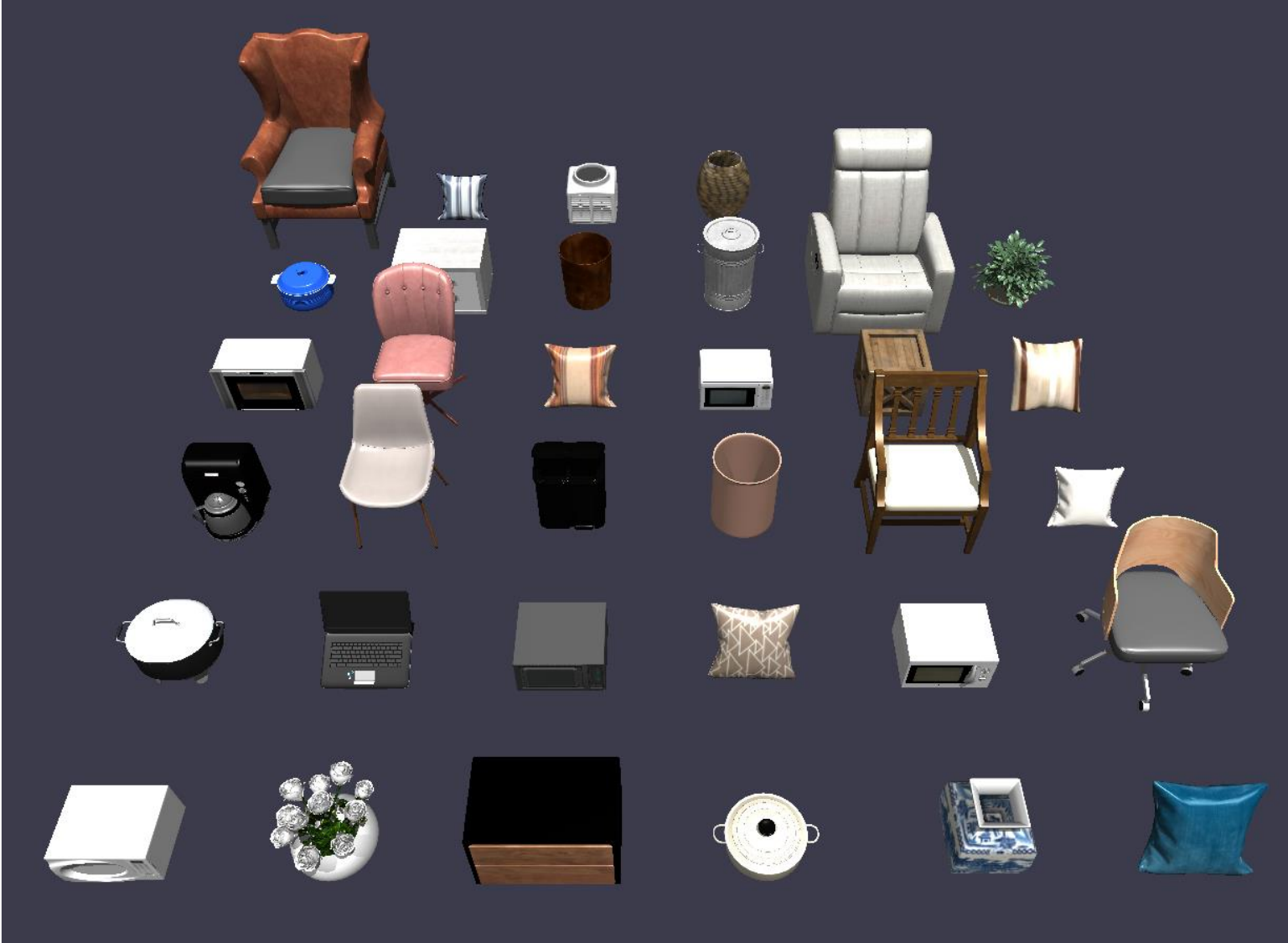}
    \caption{Movable objects in HITR dataset.}
    \label{fig:hitr-object}
\end{figure}
\subsection{Generation Process}
Figure~\ref{fig:datagen} provides an illustration of how we construct the Human-in-the-Room (HITR) dataset, following the procedural generation pipeline~\cite{procthor}.
First, we manually design four distinct room layouts: \textit{bedroom}, \textit{livingroom}, \textit{kitchen}, and \textit{warehouse}. 
These template layouts are subsequently populated with various object models to create a set of diverse scenes, which serve as the goal states for the rearrangement tasks.
To generate the initial state of each task, we randomly relocate an object to a different receptacle or to the ground. 
The initial position of the humanoid is randomly sampled from the navigable areas. 
As for the initial orientation, the humanoid heads to the object. 
Our dataset guarantees the object visibility in the first view and also covers the target position visibility in 89\% of tasks.
Notably, our dataset does not specify the initial humanoid pose; instead, it is sampled from the training motion dataset.
We then concatenate images of the goal and initial states to create a self-contained information carrier for each task. 
It is sent to a large language model, specifically \textit{gpt-4-vision}, along with language prompts, to generate the corresponding instruction.

\subsection{Statistics}
The HITR dataset contains 615 unique tasks in various rooms, examples of which are depicted in Fig.~\ref{fig:hitr-room}. 
The dataset includes 34 movable objects and 50 static objects, all of which are sourced from HSSD~\cite{hssd}. All movable objects are shown in Fig.~\ref{fig:hitr-object} and span a wide range of categories such as \textit{chair}, \textit{pillow}, \textit{plant}, \textit{coffeemaker}, among others.
To facilitate successful interaction with our humanoid model, we manually adjust the scale of object models and assign suitable weights. 
The sizes of the objects vary from 21\textit{cm} to 126\textit{cm}, and their weights range from 5\textit{kg} to 20\textit{kg}.
On average, there are 6.5 objects present in each scene.

\section{Details of the Approach}
We describe the complete details of our approach in this section.

\subsection{Training HumanVLA-Teacher}

\subsubsection{Observation Space}
\label{sec:teacher_obs}
The observation space of HumanVLA-Teacher includes proprioception, object, goal, and waypoint.
The 223-dimensional proprioception includes:\\
\begin{tabular}{*{2}{l}}
  $\bullet$ root height $\in \mathcal{R}^1$                 & $\bullet$ root rotation $\in \mathcal{R}^6$  \\ 
  $\bullet$ root linear velocity $\in \mathcal{R}^3$        & $\bullet$ root angular velocity $\in \mathcal{R}^3$\\
  $\bullet$ link position $\in \mathcal{R}^{14\times3}$          & $\bullet$ link rotation $\in \mathcal{R}^{14\times6}$  \\ 
  $\bullet$ link linear velocity $\in\mathcal{R}^{14\times3}$    & $\bullet$ link angular velocity $\in \mathcal{R}^{14\times3}$\\
\end{tabular}\\

The object state includes object position ($\mathcal{R}^{3}$), rotation ($\mathcal{R}^{6}$), linear velocity ($\mathcal{R}^{3}$), angular velocity ($\mathcal{R}^{3}$) and BPS~\cite{bps} geometry ($\mathcal{R}^{200\times3}$) encoded by delta vectors of 200 basis points.

The goal state includes the goal position ($\mathcal{R}^{3}$) and rotation ($\mathcal{R}^{6}$).

The waypoint is denoted by $x_{t}^{wp}\in\mathcal{R}^{3}$.

We follow the default AMP~\cite{amp} to use a projected observation space for the discriminator. 
They include:\\
\begin{tabular}{*{2}{l}}
  $\bullet$ root height $\in \mathcal{R}^1$                     & $\bullet$ root rotation $\in \mathcal{R}^6$  \\ 
  $\bullet$ root linear velocity $\in \mathcal{R}^3$            & $\bullet$ root angular velocity $\in \mathcal{R}^3$\\
  $\bullet$ joint rotation $\in \mathcal{R}^{12\times6}$          & $\bullet$ joint velocity $\in \mathcal{R}^{28\times1}$  \\ 
  $\bullet$ end-effector positions of left/right hand/foot, and head  $\in \mathcal{R}^{5\times3}$\\ 
  $\bullet$ object position  $\in \mathcal{R}^{3}$
\end{tabular}\\
We send 10 consecutive frames to the discriminator; thus the total dimension is $\mathcal{R}^{10\times131}$.

All these features are represented in the local coordinate of the humanoid model. Rotations are encoded using a 6-D normal-tangent representation.

\subsubsection{Action Space}
\label{sec:teacher_act}
The action space ($\mathcal{R}^{28}$) of HumanVLA-Teacher consists of the target positions for 28 Proportional-Derivative controllers. 
Predicted actions are then utilized by the controllers to generate joint torques for effective control.

\subsubsection{Network Architecture}

We adopt MLPs as the basic networks for HumanVLA-Teacher.
Each linear layer is followed by ReLU activation.
The 600-dimensional BPS feature is compressed using an MLP of [512,512,128] layers to generate a low-dimensional representation.
This representation is then concatenated with all other observations to derive the action and value. The actor, critic, and discriminator networks are all separate MLPs with hidden layers of [1024, 1024, 512].
The actor and critic networks are trained using default PPO~\cite{ppo} losses.
The discriminator is trained using a cross-entropy loss via adversarial learning.

\begin{figure}[t]
    \centering
    \includegraphics[width=.95\textwidth]{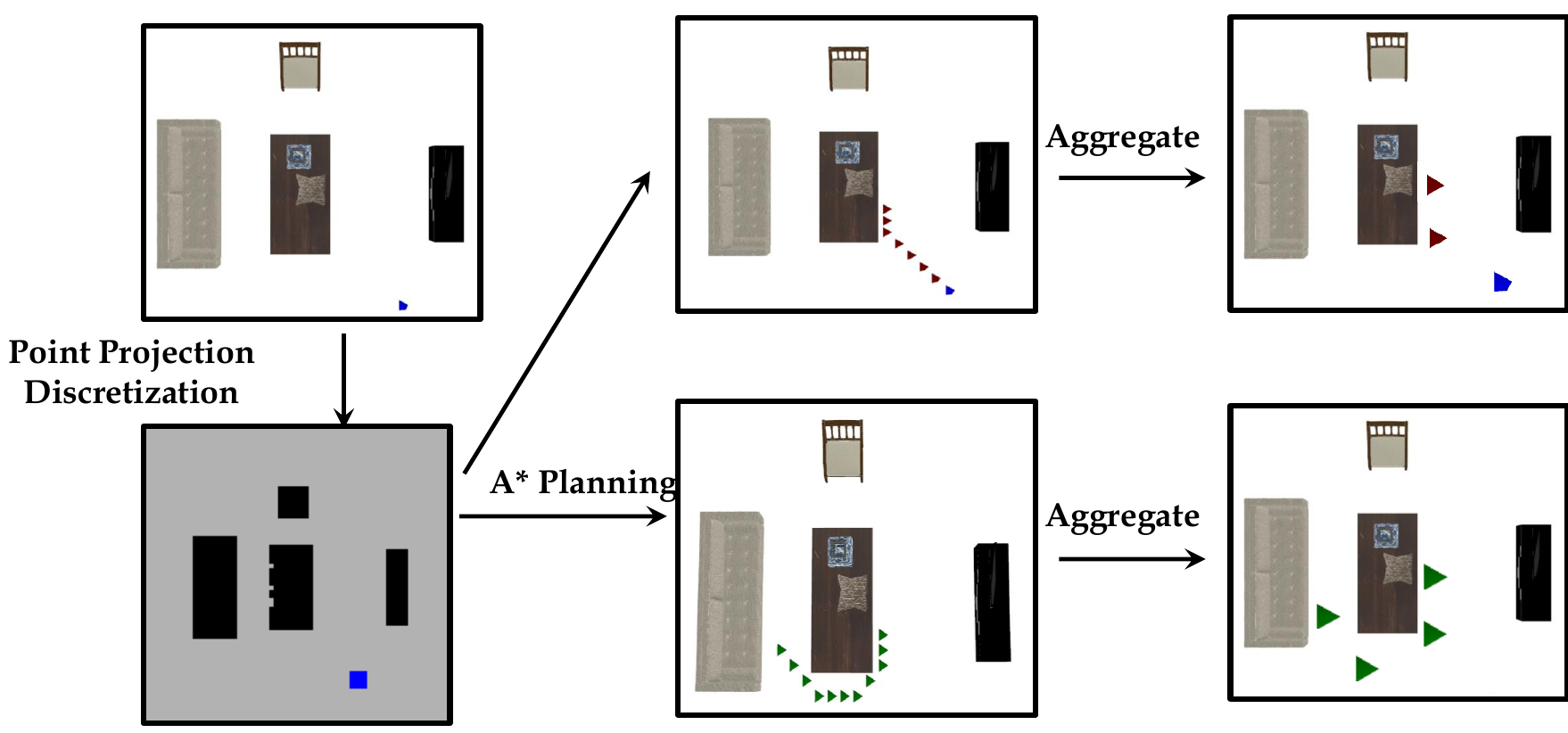}
    \caption{An overview of the path planning process. 
    The blue mark denotes the initial position. 
    Red marks denote the path from the initial position to the object. 
    Green marks denote the path from the object to the goal.}
    \label{fig:pathplan}
\end{figure}

\subsubsection{In-context Path Planning}
Navigating through intricate scenes is challenging due to the potential for unexpected object collisions. 
To mitigate this, we employ in-context path planning to facilitate collision-free locomotion, as depicted in Fig.~\ref{fig:pathplan}. 
This process involves two substreams: planning a path from the humanoid to the object, and then from the object to the goal.
Initially, we sample point clouds from all objects in the scene. These points are then projected top-down onto the ground and divided into 50\textit{cm} x 50\textit{cm} grids to construct a navigation map. 
We utilize the $A^*$ algorithm~\cite{hart1968formal} to plan two paths: one from the humanoid to the object, and another from the object to the goal.
Each path generated by the $A^*$ algorithm is represented by a series of densely packed waypoints. 
We consolidate waypoints that share a consistent moving direction to create a sparse waypoint set. 
During task execution, the humanoid model is guided by a sequence of these waypoints and is encouraged to move toward the waypoint at each step.
Once the humanoid reaches the waypoint within a 50\textit{cm} distance, the waypoint proceeds to the next.
The last waypoint is the goal position.
The waypoint is used in reward computation to guide the movement, described in the following sections.

\subsubsection{Carry Curriculum Pre-training}

We uniformly conceptualize object rearrangement as a three-step process: locomotion towards the object, making contact with the object, and then relocating the object to achieve the goal. 
However, directly training these three processes from scratch can be exceptionally challenging.
The main difficulty lies in the fact that physical interactions with objects necessitate contacting prior, such as robust object holding, to enable subsequent object dynamics. 
Through empirical experiments, we have found that carrying objects to other receptacles is significantly more difficult than pushing and pulling objects on the ground. 
This underscores the need to learn a carry prior to boosting the subsequent object relocation.

Inspired by the curriculum learning paradigm~\cite{bengio2009curriculum}, we design a carry curriculum pre-training scheme to learn the carry prior. 
It encompasses the initial two steps, namely locomotion and contacting, which form an easier curriculum compared to the difficult three-step rearrangement task.
We exclude on-ground objects in carry curriculum pre-training, whose initial height and goal height are both smaller than 0.1\textit{m}.
These objects, such as \textit{small table}, \textit{chair}, can be relocated by direct on-ground pushing and pulling without a lift.
The goal of the carry curriculum is to enable the agent to walk towards the object and establish a robust contact, allowing it to securely hold the object in the air; thus, we define reward as a combination of walking reward and contacting reward:
\begin{gather}
    r_t = r_t^{walk} + r_t^{contact}
\end{gather}
where $r^{walk}_t$ encourages the humanoid to walk to the object. 
Specifically, it encourages a closer distance between root position $x_{t}^{root}$ and object position $x_{t}^{root}$, as well as walking to the waypoint direction $x_t^{wp}$ at a target speed $v^*$=1.5\textit{m/s}:
\begin{gather}
    r_t^{walk} = 
    \begin{cases}
        0.1\exp(-0.5|| x_{t}^{obj} - x_{t}^{root}||) + 0.2\exp(-2||v^* - v_t^{root}\cdot x_t^{wp}||^2) & || x_{t}^{obj} - x_{t}^{root}|| > 0.5\\
        0.3 & \text{otherwise}
    \end{cases}
\end{gather}
$r_t^{contact}$ encourages the humanoid hands $x_{t}^{hand}$ to contact with the object $x_{t}^{obj}$, followed by lifting the object up for $\Delta h = 0.3m$ from an initial height $h_{init}^{obj}$. It is formulated as:
\begin{gather}
    r_t^{contact} = 
        0.2\exp(-10|| x_{t}^{obj} - x_{t}^{hand}||) + 0.5~\text{clip}(h_{t}^{obj} - h_{init}^{obj},0,\Delta h)/\Delta h 
\end{gather}

\subsubsection{Rearrangement Learning}
The complete rearrangement task consists of the whole three-step process. 
The reward is formulated as a combination of walking reward, contacting reward, and relocation reward:
\begin{equation}
    r_t = r_t^{walk} + r_t^{contact} + r_t^{relocation}
\end{equation}
where $r_t^{walk}$, $r_t^{contact}$ have similar formulation but different weights and conditions compared to the carry curriculum pre-training, specifically:
\begin{gather}
    r_t^{walk} = 
    \begin{cases}
        0.1\exp(-0.5|| x_{t}^{obj} - x_{t}^{root}||)+0.1\exp(-2||v^* - v_t^{root}\cdot x_t^{wp}||^2) & ||x_{t}^{obj}-x_{t}^{root}||>0.5 \\
        0.2 & \text{otherwise}
    \end{cases}\\
     r_t^{contact} = \begin{cases}
        0.1\exp(-10|| x_{t}^{obj} - x_{t}^{hand}||) + 0.1~\text{clip}(h_{t}^{obj} - h_{init}^{obj},0,\Delta h)/\Delta h & ||x_{t}^{obj}-x_{goal}^{obj}||>0.5\\
        0.2 & \text{otherwise}
        \end{cases}
\end{gather}
In the context of the rearrangement task, we do not set an explicit lifting condition, but instead require the contact prior to enhance subsequent dynamics. 
As a result, the height change, $\Delta h$ is reduced to 0.1\textit{m} for objects that are going to be carried, and to zero for objects that are consistently on the ground.

$r_t^{relocation}$ comprises four components, namely $r_t^{vel}$, $r_t^{far}$, $r_t^{near}$, and $r_t^{rot}$. 
$r_t^{vel}$ encourages moving the object to the next waypoint $x_t^{wp}$.
$r_t^{far}$ encourages a close distance to the next waypoint.
$r_t^{near}$ encourages to meet the goal position $x_{goal}$.
$r_t^{rot}$ encourages to meet the rotation position $q_{goal}$.
They are formulated as:
\begin{gather}
    r_t^{relocation} = 0.1r_t^{vel}+0.2r_t^{far} + 0.2r_t^{near} + 0.1r_t^{rot}\\
    r_t^{vel} = 
    \begin{cases}
        \exp(-2||v^* - v_t^{obj}\cdot x_t^{wp}||^2) & ||x_{t}^{obj}-x_{goal}^{obj}||>0.5\\
        1 & \text{otherwise}
    \end{cases}\\
    r_t^{far} = \exp(-||x_{t}^{obj} - x_{t}^{wp}||)\\
    r_t^{near} = \exp(-5||x_{t}^{obj} - x_{goal}||)\\
    r_t^{rot} = \exp(-2||q_{t}^{obj} - q_{goal}||)
\end{gather}

\subsubsection{Hyperparameter Setting}
We provide a hyperparameter table for HumanVLA-Teacher training in Tab.~\ref{tab:param-teacher}. 
It is shared for both the carry curriculum pre-training and rearrangement learning.

\begin{table}[!t]
\caption{Hyperparameters for HumanVLA-Teacher training.}
\label{tab:param-teacher}
\centering
\resizebox{0.99 \linewidth}{!}{
\begin{tabular}{cccccc}
\toprule
Hyperparameter & Value & Hyperparameter & Value & Hyperparameter & Value\\
\midrule
Num. envs & 16,384 & Max episode length & 300 & Num. epochs & 30,000\\
Discount factor & 0.99 & GAE parameter & 0.95 & Observation clip & 5 \\
Action clip & 1 & Optimizer & Adam & Learning rate & 3e-5 \\
Actor loss weight & 1 & Critic loss weight & 5 & Discriminator loss weight & 5 \\
Num. rollouts per PPO update & 32 & PPO clip & 0.2 & PPO miniepochs & 5 \\
Num. batches per miniepoch & 8 & AMP consecutive frames $t^*$ & 10 & Gradient penalty $w^{gp}$ & 5 \\
Task reward weight $w^G$ & 1 & Style reward weight $w^S$ & 1 & Min style clipping bound $\xi_{min}$ & 0.4 \\
\bottomrule
\end{tabular}

}
\end{table}

\begin{table}[!t]
\caption{Hyperparameters for HumanVLA training.}
\label{tab:param-student}
\centering
\resizebox{0.99 \linewidth}{!}{
\begin{tabular}{cccccc}
\toprule
Hyperparameter & Value & Hyperparameter & Value & Hyperparameter & Value\\
\midrule
Num. envs & 585 & Max episode length & 300 & Num. epochs & 20,000\\
Observation clip & 5 & Action clip & 1 & Optimizer & Adam \\
Learning rate & 5e-4 & Num. rollouts per epoch & 1  & Num. train steps per epoch & 5\\
Batch size & 600 & DAgger $\beta_0$ & 1 & DAgger $\lambda$ & 0.998 \\ 
Camera resolution & 256,256 & Camera FoV & 90 &Active rendering $w^{AR}$ & 0.2 \\

\bottomrule
\end{tabular}

}
\vspace{-0.3cm}
\end{table}

\subsection{Training HumanVLA}

\subsubsection{Observation Space}
The observation space of HumanVLA includes proprioception, last action, egocentric image, and language instruction.

The proprioception adheres to the 223-dimensional feature defined in the HumanVLA-Teacher (Sec.~\ref{sec:teacher_obs}).

The last action is denoted by $a_{t-1}\in\mathcal{R}^{28}$. An all-zero feature is used for it at the first step.

The egocentric image has 256 x 256 pixels with a field of view spanning 90 degrees.
We mount a camera on the head of the humanoid model, with the camera position offset from the head being [0.103, 0, 0.175].
The camera direction aligns with the forward direction of the head.

The natural language instruction specifies the task goal.

\subsubsection{Action Space}
The action space for HumanVLA aligns with the 28-dimensional HumanVLA-Teacher action space defined in Sec.~\ref{sec:teacher_act}.

\subsubsection{Network Architecture}
The HumanVLA network consists of an image encoder, a text encoder, and an action decoder.
The image encoder is an \textit{EfficientNet-B0}~\cite{efficientnet} while the text encoder is a frozen \textit{bert-base-uncased}~\cite{bert}.
Our primary design choices for these two models are their fast inference speed.
The 1280-dimensional \textit{EfficientNet-B0} feature is passed through a linear layer to yield a compressed 128-dimensional feature.
An MLP with [512,128] hidden layers is used to compress the 768-dimensional \textit{bert-base-uncased} feature down 128 dimensions.
The compressed image and text features are then concatenated with the proprioception and last action, and this combined feature is sent into the action decoder.
The action decoder is a 6-layer MLP,  with each hidden layer being 1024-dimensional. 
Each linear layer is followed by BatchNorm and ReLU. 
Skip connections are used between the first and third layers, as well as between the third and fifth layers.

\subsubsection{Active Rendering Action}
We propose the active rendering technique to enhance the quality of visual perception and facilitate object rearrangement. 
In this section, we provide more details on how to compute the active rendering action, specifically applied to the neck to adjust the camera pose.
The primary regulation focuses on the camera's target view, which is the forward direction. 
To determine this, we compute the center of the object point cloud and use it as the target viewpoint. 
The direction from the camera position to the point cloud center represents the expected forward direction of both the camera and the head.
Since the neck joint is a 3-DoF spherical joint, a single regulation can lead to ambiguous actions. 
To address this, we introduce the second regulation that controls the side direction of the head. 
It is perpendicular to the plane formed by the upward torso and the camera view. 
These two regulations result in a unique head rotation.
Finally, inverse kinematics is used to solve neck actions.

\subsubsection{Learning Process}
We train HumanVLA by cloning actions from HumanVLA-Teacher.
At each step, the HumanVLA-Teacher predicts an action using privileged information.
It is mixed with an active rendering action to yield the label for supervision.
HumanVLA is optimized to minimize the mean square error.

We adopt a DAgger~\cite{dagger} framework to manage the online learning process.
DAgger iteratively schedules a mixed policy $\pi = \beta_{t} \pi^{tch} + (1-\beta_t) \pi^{stu}$ at epoch $t$ to explore the environment and expand a training dataset.
It can alleviate the covariate shift problem between different policies.
We use an exponential function to schedule the mixing factor $\beta_t = \beta_0*\lambda^t$.


\subsubsection{Hyperparameter Setting}
We provide a hyperparameter table for HumanVLA training in Tab.~\ref{tab:param-student}. 
\begin{figure}[!t]
  \centering
  \begin{minipage}[t]{0.46\textwidth}
    \centering
    \includegraphics[width=\textwidth]{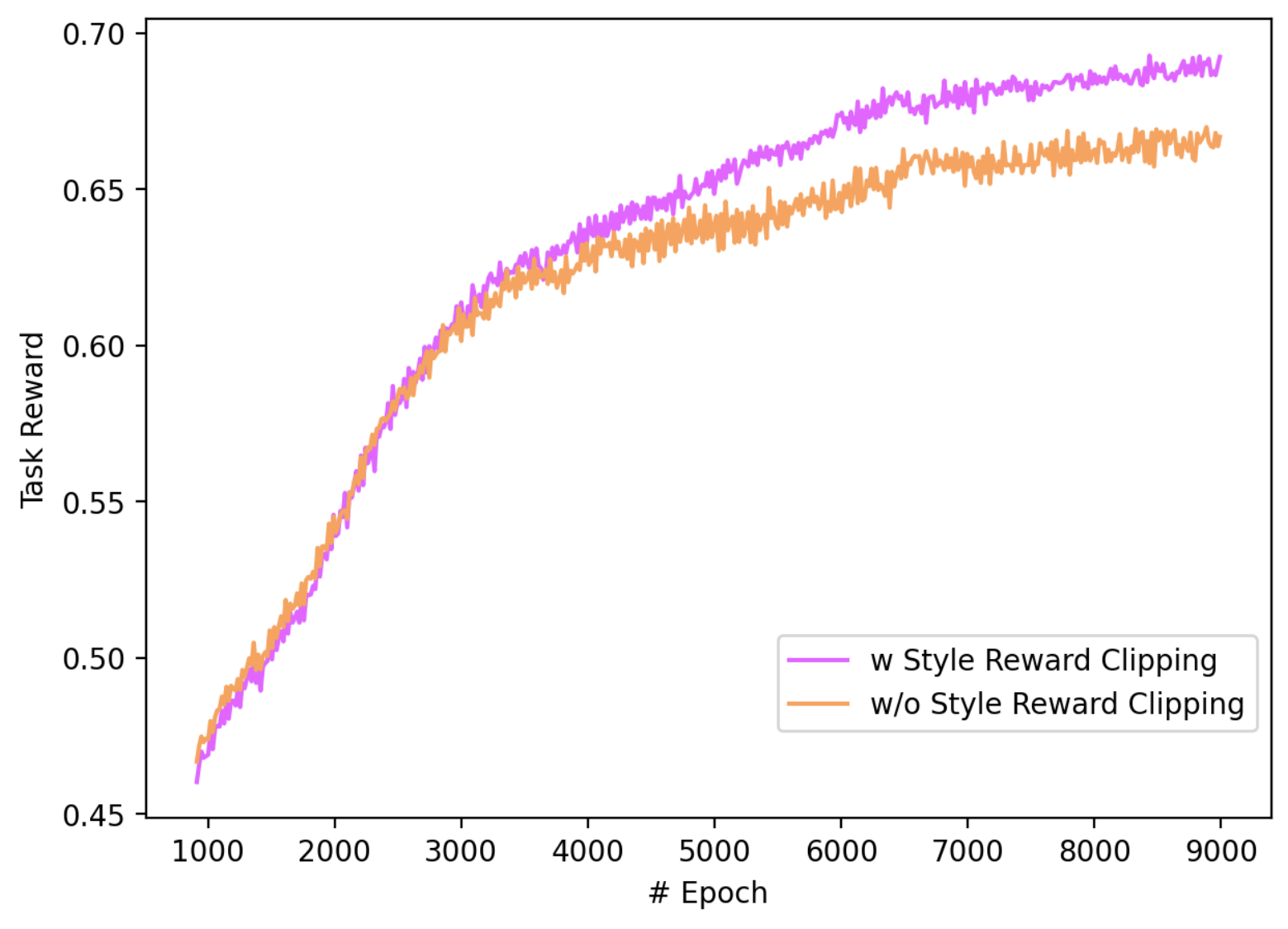}
    \caption{Learning curve comparison w/ and w/o style reward clipping.}
    \label{fig:curve_src}
  \end{minipage}
  \hfill
  \begin{minipage}[t]{0.5\textwidth}
    \centering
    \includegraphics[width=\textwidth]{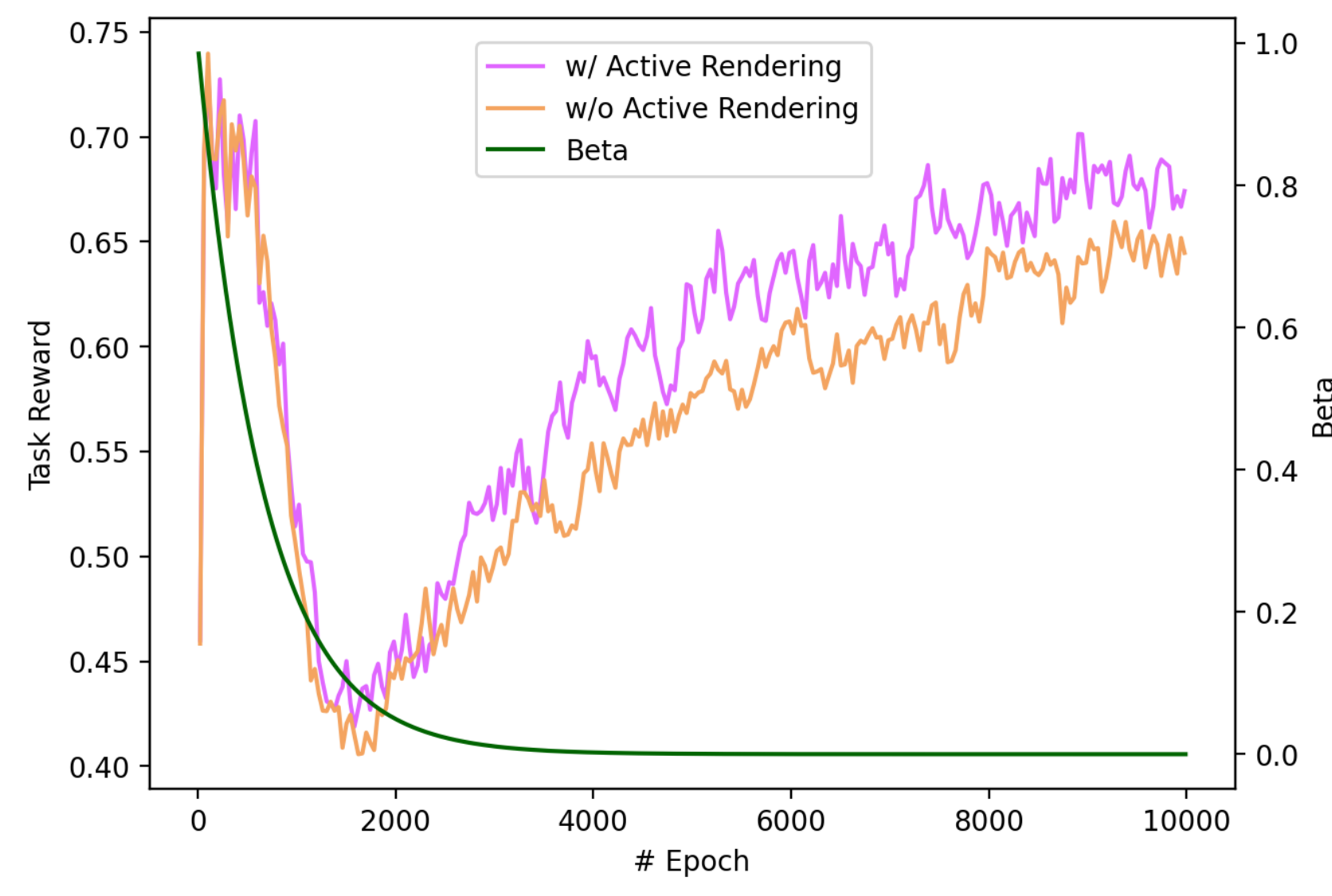}
    \caption{
    Learning curve comparison w/ and w/o active rendering.
    The process is dominated by the teacher policy in the early stage with high $\beta$ and demonstrates the reward upper bound.
    }
    \label{fig:curve_ar}
  \end{minipage}
  \vspace{-0.3cm}
  \label{fig:combined}
\end{figure}

\section{Additional Results}

\begin{figure}[t]
    \centering
    \includegraphics[width=0.8\textwidth]{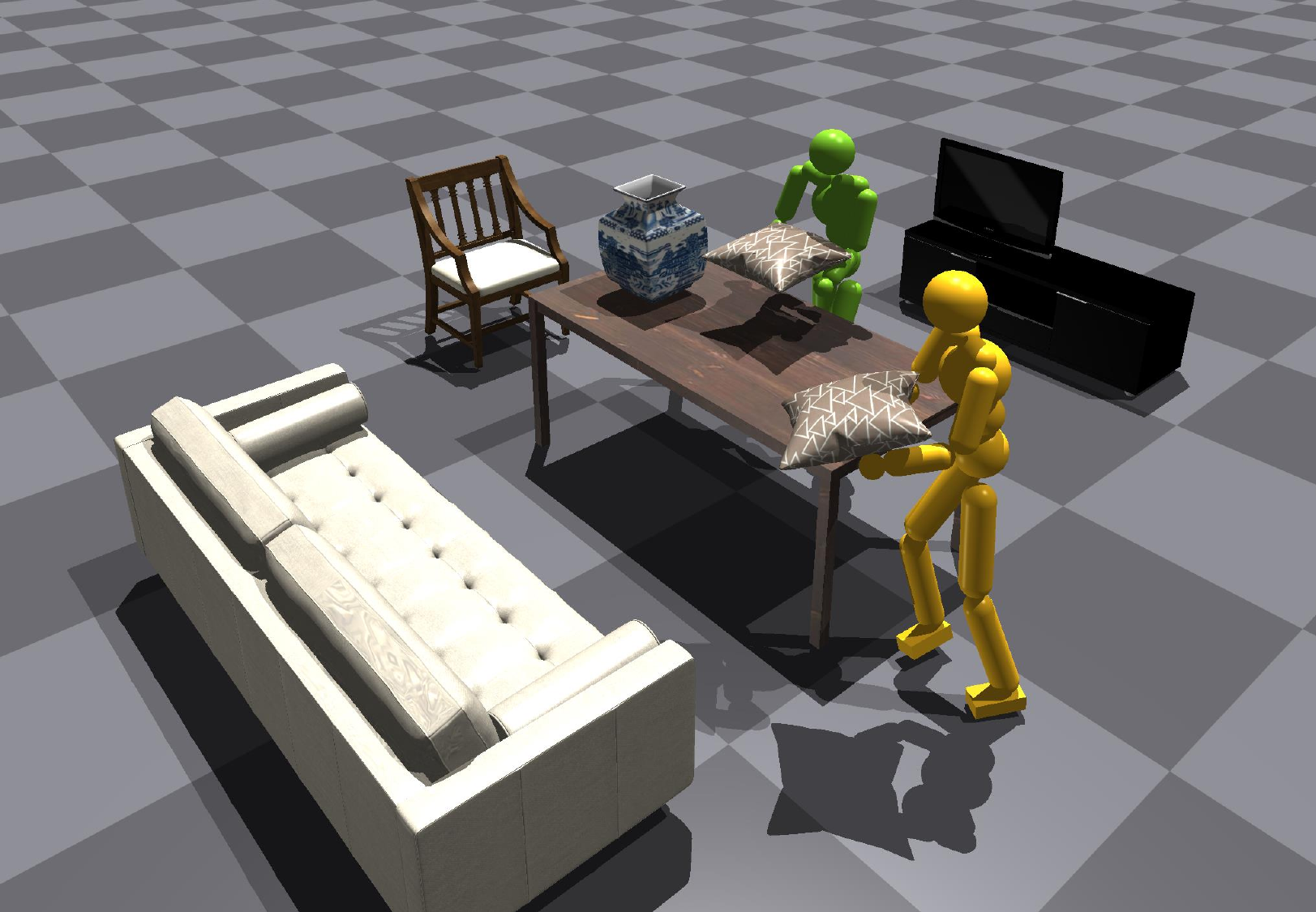}
    \caption{
    Comparison w/ and w/o path planning.
    The green humanoid without path guidance fails to get close to the sofa, while the yellow humanoid with path guidance learns to go around the central table. 
    Instruction: Move the pillow to the sofa.
    }
    \label{fig:vis_path}
\end{figure}

\subsection{Generalization}

The primary test set of HITR contains task-level unseen tasks including new compositions of objects in the scene, new placement of objects, and regenerated new text instructions from LLM describing new compositions and new spatial relations. We have proven the generalization ability of HumanVLA in unseen tasks.

However, generalization in any unseen without any similar pattern in the seen data is super challenging, which is also an ultimate goal of embodied AI research. 
We build extra tiny unseen data to make additional analysis and further disclose our method. We make additional testing data: (1) Unseen texts generated for training tasks, manually reviewed to be distinct from training data. (2) Unseen objects by changing visual appearance in training tasks. (3) Unseen object category (cup) with different geometry. (4) Unseen scene layouts by repositioning static large objects.

Results are reported in Tab.~\ref{tab:unseen_append}. We find that our work suffers less from unseen texts and unseen visual appearance. But generalizing to unseen object categories and execution in the unseen scenes remains a main challenge.

\begin{table}[!t]
\caption{Unseen data analysis.}
\label{tab:unseen_append}
\centering
\resizebox{0.9 \linewidth}{!}{

\begin{tabular}{l|ccc}
\toprule
& Success Rate (\%) $\uparrow$ & Precision (\textit{cm})$\downarrow$ & Execution Time (\textit{s}) $\downarrow$\\
\midrule
Useen Text	             &   65	 &   50.4    &	5.4 \\
Unseen object (visual)	 &   50	 &   72.3	 &  6.2 \\ 
Unseen object (geometry) &   20	 &  118.8	 &  7.9 \\ 
Unseen scene layout	     &   35  &	 88.5	 & 6.8 \\
\bottomrule
\end{tabular}

}
\end{table}

\begin{figure}[t]
    \centering
    \includegraphics[width=0.9\textwidth]{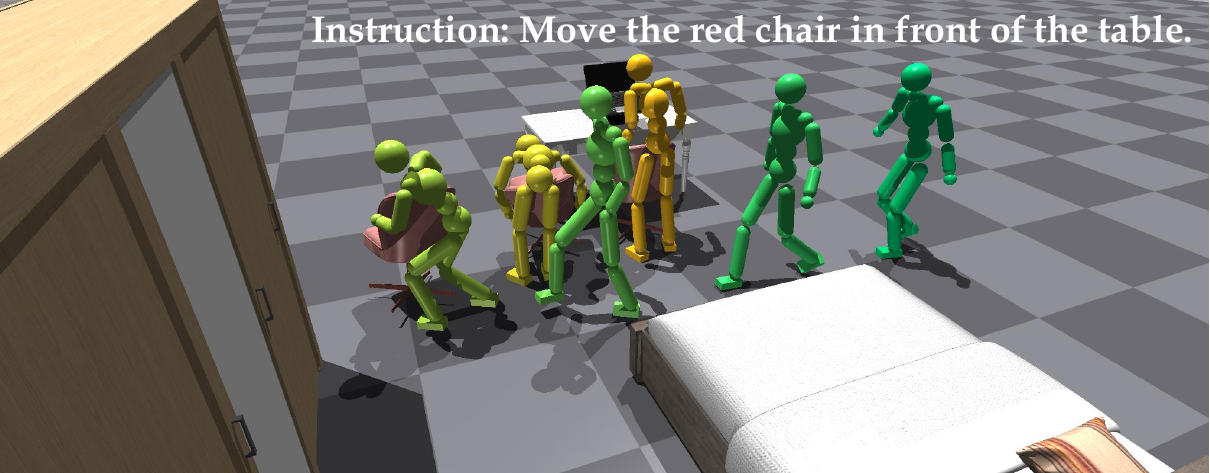}
    \includegraphics[width=0.9\textwidth]{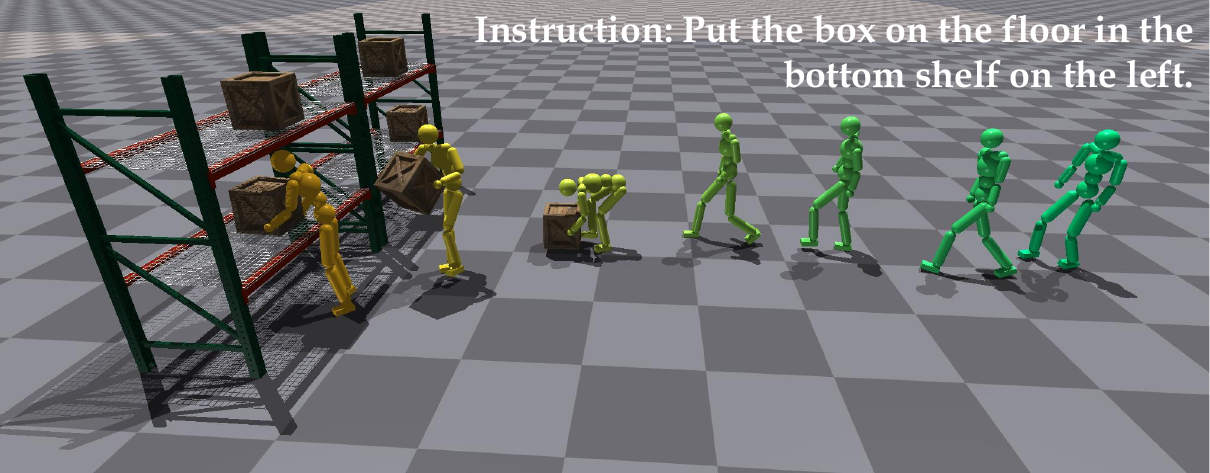}
    \caption{Additional qualitative results.}
    \label{fig:qualitative_supp}
    \vspace{-0.3cm}
\end{figure}

\subsection{Learning Curves}
We plot the learning curves in Fig.~\ref{fig:curve_src} and Fig.~\ref{fig:curve_ar} to demonstrate the efficacy of our method.
In Fig.~\ref{fig:curve_src}, our method has been proven to converge faster in task completion with style reward clipping.
In Fig.~\ref{fig:curve_ar}, active rendering improves perception quality and facilitates the learning of the student policy.

\subsection{Qualitative Ablation}
A qualitative ablation about the path planning module is represented in Fig.~\ref{fig:vis_path}. 
Specifically, the green humanoid fails to navigate close to the goal receptacle, \ie, the 
sofa.
However, the yellow humanoid is guided by planned waypoints to go around the center table and place the pillow on the sofa.

\subsection{Additional Qualitative Results}
We provide additional qualitative visualizations in Fig.~\ref{fig:qualitative_supp} to disclose our results.
\newpage


\end{document}